# Development of a multi-user handwriting recognition system using Tesseract open source OCR engine

**Sandip Rakshit** [1], **Subhadip Basu** [2] [#]

[1] *Techno India College of Technology, Kolkata, India*
[2] *Computer Science and Engineering Department, Jadavpur University, India*

[#] *Corresponding author. E-mail: subhadip@ieee.org*

## Abstract

*The objective of the paper is to recognize handwritten samples of lower case Roman script using Tesseract open source Optical Character Recognition (OCR) engine under Apache License 2.0. Handwritten data samples containing isolated and free-flow text were collected from different users. Tesseract is trained with user-specific data samples of both the categories of document pages to generate separate user-models representing a unique language-set. Each such language-set recognizes isolated and free-flow handwritten test samples collected from the designated user. On a three user model, the system is trained with 1844, 1535 and 1113 isolated handwritten character samples collected from three different users and the performance is tested on 1133, 1186 and 1204 character samples, collected form the test sets of the three users respectively. The user specific character level accuracies were obtained as 87.92%, 81.53% and 65.71% respectively. The overall character-level accuracy of the system is observed as 78.39%. The system fails to segment 10.96% characters and erroneously classifies 10.65% characters on the overall dataset.*

Keywords: handwriting recognition, isolated roman characters, Tesseract, OCR engine

## 1. Introduction

*Optical Character Recognition (OCR)* systems ease the barrier of the keyboard interface between man & machine to a great extent, and help in office automation with huge saving of time and human effort. Such systems allow desired manipulation of the scanned text as the output is coded with ASCII or some other character code from the paper based input text. For a specific language based on some alphabet, OCR techniques are either aimed at printed text or handwritten text. The present work is aimed at the later.

Machine recognition of handwritten text is one of the challenging areas of research for the pattern recognition community. In general, OCR systems have potential applications in extracting data from filled in forms, interpreting handwritten addresses from postal documents for automatic routing, automatic reading of bank cheques etc. The core component of such application software is an OCR engine, equipped with the key functional modules like line extraction, line-to-word segmentation, word-to-character segmentation, character recognition and word-level lexicon analysis using standard dictionaries.

Development of a handwritten OCR engine with high recognition accuracy is a still an open problem for the research community. Lot of research efforts have already been reported [1-8] on different key aspects of handwritten character recognition systems. In the current work, instead of developing a new handwritten OCR engine from scratch, we have used Tesseract 2.01 [9], an open source OCR Engine under Apache License 2.0, for recognition of handwritten pages consisting of lower case characters of Roman script. Tesseract OCR engine provides high level of character recognition accuracy on poorly printed or poorly copied dense text. In one of our earlier works [10], we had developed a system for estimation of recognition accuracy of Tesseract OCR engine on handwritten lower case character samples, collected from a single user. But the performance of this OCR engine could not be tested extensively on handwriting samples of multiple users. This has been one of the major motivations behind the current work, presented in this paper.

In the current work, we have used Tesseract to perform user specific training on handwriting samples of both isolated and free-flow texts, written using lower case Roman script. The performance is evaluated on both the categories of document pages for observation of character level and word level accuracies.

## 2. Overview of the Tesseract OCR engine

Tesseract is an open source (under Apache License 2.0) offline optical character recognition engine, originally developed at Hewlett Packard from 1984 to 1994. Tesseract was first started as a PhD research project in HPLabs, Bristol [11]. In the year 1995 it is sent to UNLV where it proved its worth against the commercial engines of the time [12]. In the year 2005 Hewlett Packard and University of Nevada, Las Vegas, released it. Now it is partially funded by Google [13] and released under the Apache license, version 2.0. The latest version, Tesseract 2.03 is released in April, 2008. In the current work, we have used Tesseract version 2.01, released in August 2007.

Like any standard OCR engine, Tesseract is developed on top of the key functional modules like, line and word finder, word recognizer, static character classifier, linguistic analyzer and an adaptive classifier. However, it does not support document layout analysis, output formatting and graphical user interface. Currently, Tesseract can recognize printed text written in English, Spanish, French, Italian, Dutch, German and various other languages.

To train Tesseract in English language 8 data files are required in tessdata sub directory. The 8 files used for English are to be generated as follows:

    tessdata/eng.freq-dawg
    tessdata/eng.word-dawg
    tessdata/eng.user-words
    tessdata/eng.inttemp
    tessdata/eng.normproto
    tessdata/eng.pffmtable
    tessdata/eng.unicharset
    tessdata/eng.DangAmbigs

## 3. The present work

In the current work, Tesseract 2.01 is used for recognition of handwriting samples of both isolated and free-flow texts, written using lower case Roman script. Key functional modules of the developed system are discussed the following sub-sections.

### 3.1. Collection of the dataset

For collection of the dataset for the current experiment, we have concentrated on lower case characters of Roman script. Six handwritten document pages were collected from each of the three different users in two types of datasets. In the first dataset (Datasaet-1), four pages of isolated handwritten lower case Roman characters were collected, as shown in Fig. 1(a), and in the second dataset (Datasaet-2), two pages of free-flow handwritten text, as shown in Fig. 1(b), written from technical articles, were collected from each user. For each user, three pages from the first set and one page from the second dataset were considered for training the Tesseract OCR engine. The remaining two pages, one from each set, constitute the test set for the current experiment.

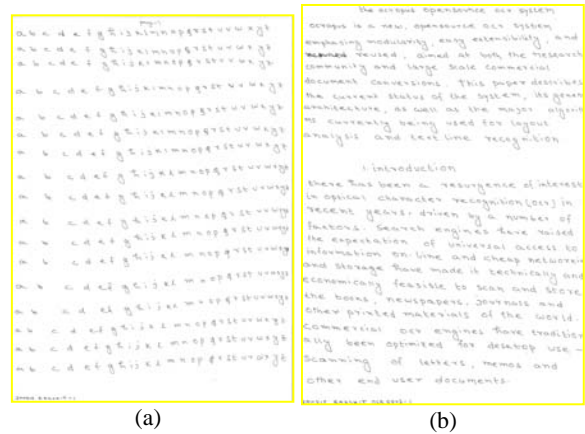

(a)　　　　　　(b)

**Fig. 1(a-b). Sample document pages containing training sets of isolated characters and free flow text**

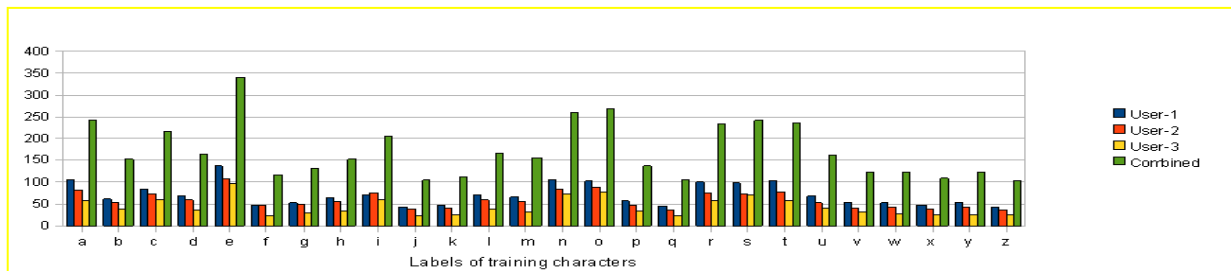

**Fig. 2. Frequency distribution of different character samples during training**

The training dataset contains around 70 sample sets of isolated lower case Roman characters for each user and around 120 words (around 650 characters) of free-flow text. For example, the frequency of different lower case characters in the training set for different users, and the overall training character frequency is shown in Fig. 2.

**3.2. Labeling training data**

For labeling the training samples using Tesseract we have taken help of a tool named bbTesseract [13]. To generate the training files for a specific user, we need to prepare the box files for each training images using the following command:

*tesseract fontfile.tif fontfile batch.nochop makebox*

The box file is a text file that includes the characters in the training image, in order, one per line, with the coordinates of the bounding box around the image. The new Tesseract 2.01 has a mode in which it will output a text file of the required format. Some times the character set is different to its current training, it will naturally have the text incorrect. In that case we have to manually edit the file (using bbTesseract) to correct the incorrect characters in it. Then we have to rename fontfile.txt to fontfile.box. Fig. 3 shows a screenshot of the bbTesseract tool, used for labeling the training set.

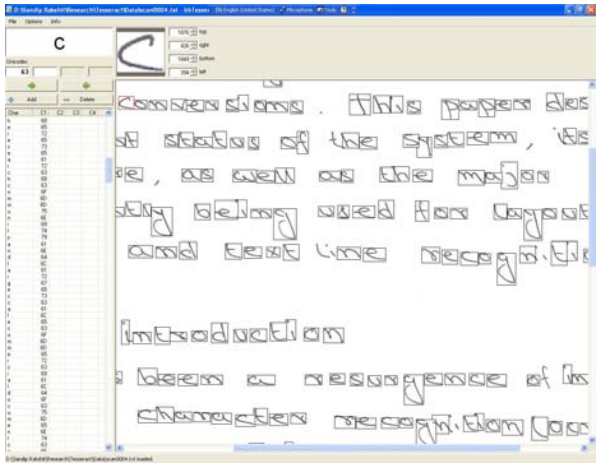

**Fig.3. A sample screenshot of the bbTesseract tool**

**3.3. Training the data using Tesseract OCR engine**

For training a new handwritten character set for any user, we have to put in the effort to get one good box file for a handwritten document page, run the rest of the training process, discussed below, to create a new language set. Then use Tesseract again using the newly created language set to label the rest of the box files corresponding to the remaining training images using the process discussed in section 3.2.

For each of our training image, boxfile pairs, run Tesseract in training mode using the following command:

*tesseract fontfile.tif junk nobatch box.train*

The output of this step is fontfile.tr which contains the features of each character of the training page. The character shape features can be clustered using the mftraining and cntraining programs:

*mftraining fontfile_1.tr fontfile_2.tr ...*

This will output three data files: inttemp , pffmtable and Microfeat, and the following command:

*cntraining fontfile_1.tr fontfile_2.tr ...*

This will output the normproto data file. Now, to generate the unicharset data file, unicharset_extractor program is used as follows:

*unicharset_extractor fontfile_1.box fontfile_2.box ...*

Tesseract uses 3 dictionary files for each language. Two of the files are coded as a Directed Acyclic Word Graph (DAWG), and the other is a plain UTF-8 text file. To make the DAWG dictionary files a wordlist is required for our language. The wordlist is formatted as a UTF-8 text file with one word per line. The corresponding command is:

*wordlist2dawg frequent_words_list freq-dawg*
*wordlist2dawg words_list word-dawg*

The third dictionary file name is user-words and is usually empty. The final data file of Tesseract is DangAmbigs file. This file cannot be used to translate characters from one set to another. The DangAmbigs file may be empty also.

Now we have to collect all the 8 files and rename them with a lang. prefix, where lang is the 3-letter code for our language and put them in our tessdata directory. Tesseract can then recognize text in our language using the command:

*tesseract image.tif output -l lang*

## 4. Experimental results

For conducting the current experiment, we have considered a three user model for preparing three different language sets using Tesseract open source OCR engine. As discussed in section 3.1, the training and test patterns of each individual user is spread over two types of data collection sheets, viz., isolated text and free-flow text. The overall distribution of training and test samples of the three different users is shown in Table 1. The sample test pages used for this experiment, for any user, are shown in Fig. 4. The experiment was focused on testing the core recognition accuracy of Tesseract OCR engine on handwritten document pages. For this purpose, the linguistic analysis module of Tesseract, involving the language files freq-dawg, word-dawg, user-words and DangAmbigs are purposefully left blank.

**Table 1. Distribution of training and test samples of different users**

|  | Isolated Characters | Free flow Text | | Total Characters |
|---|---|---|---|---|
|  |  | Characters | Words |  |
| Train set for User-1 | 1185 | 659 | 137 | 1844 |
| Test set for User-1 | 442 | 691 | 120 | 1133 |
| Train set for User-2 | 1006 | 529 | 130 | 1535 |
| Test set for user-2 | 468 | 718 | 128 | 1186 |
| Training set for user-3 | 588 | 525 | 169 | 1113 |
| Test Set for user-3 | 260 | 944 | 161 | 1204 |

The performance of the developed system is evaluated on the two datasets for each of the users, as discussed in section 3.1. To evaluate the performance of the present technique the following expression is developed.

Recognition accuracy = $(C_t / (C_m + C_s))*100$

Where $C_t$ = the number of character segments producing true classification result and $C_m$ = the number of misclassified character segments and $C_s$ signifies the number of character Tesseract fails to segment, i.e., producing under segmentation. The rejected character/word samples are excluded from computation of recognition accuracy of the designed system.

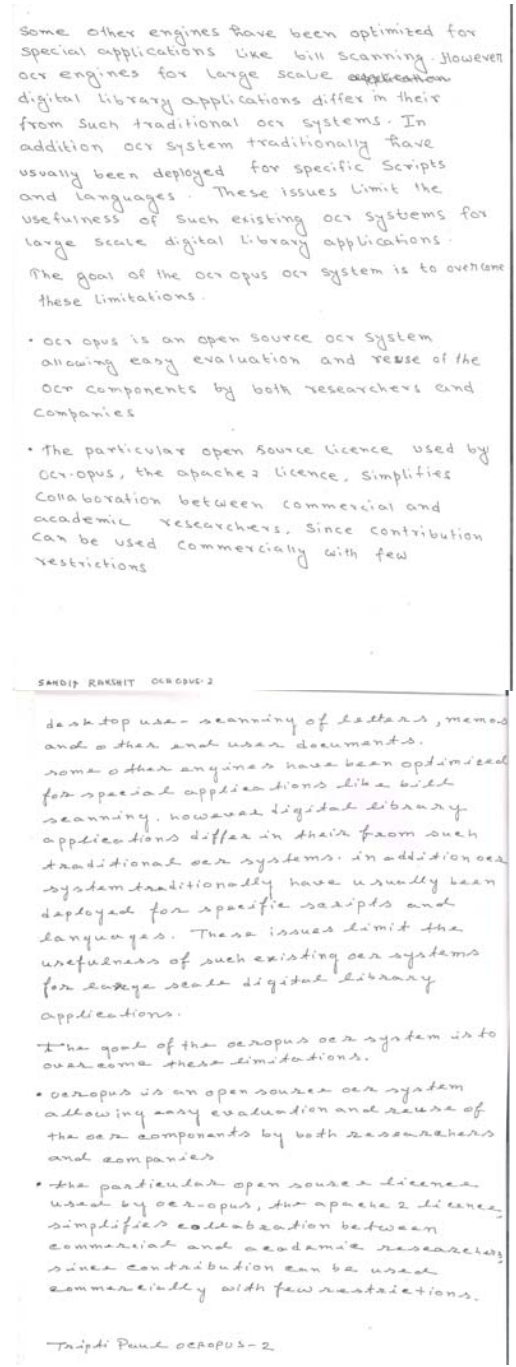

**Fig.4. Snapshots of the test pages used for the current experiment**

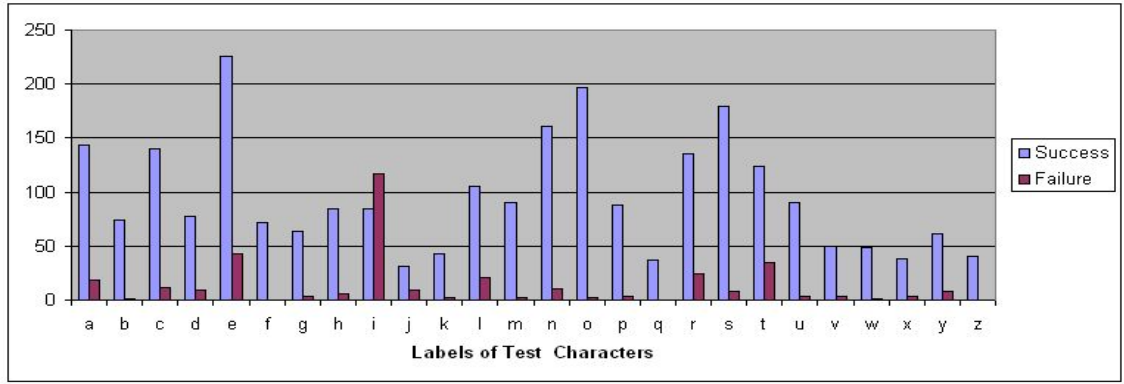

**Fig. 5. Distribution of success and failure cases over the free flow test page.**

Table 2(a-c) shows an analysis of successful classification, misclassification, segmentation failure and rejection results on the test samples of the three users. Fig. 5 shows a character wise distribution of success and failure accuracies on the overall test dataset. As observed from the experimentation a significant proportion rejection cases evolve out of the word segmentation failures. This is so because Tesseract is originally designed to recognize printed document pages with uniformity in baseline and character/word spacings. Another source of error is due to the internal segmentation of some of the characters. More specifically, the character 'i' often gets internally segmented into two parts, leading to high individual error rates.

**Table 2. Analysis of recognition performance of the developed system**
**(a) Recognition performance of User-1 test dataset**

|  | Characters of Dataset-1 | Characters of Dataset-2 | Overall performance |
|---|---|---|---|
| Successful Recognition | 95.42 | 83.2 | 87.92 |
| Misclassification | 4.1 | 16.19 | 11.52 |
| Segmentation Failure | 0.48 | 0.61 | 0.56 |
| Rejection | 6.10 | 4.34 | 5.03 |

**(b) Recognition performance of User-2 test dataset**

|  | Characters of Dataset-1 | Characters of Dataset-2 | Overall performance |
|---|---|---|---|
| Successful Recognition | 91.62 | 76.45 | 81.53 |
| Misclassification | 8.38 | 18.31 | 15.00 |
| Segmentation Failure | 0.00 | 5.24 | 3.47 |
| Rejection | 26.07 | 4.18 | 12.82 |

**(c) Recognition performance of User-3 test dataset**

|  | Characters of Dataset-1 | Characters of Dataset-2 | Overall performance |
|---|---|---|---|
| Successful Recognition | 92.34 | 58.35 | 65.71 |
| Misclassification | 6.81 | 6.24 | 6.36 |
| Segmentation Failure | 0.85 | 35.41 | 27.93 |
| Rejection | 9.61 | 9.96 | 9.88 |

As shown in Table 2(a-c), the overall character-level recognition accuracy of the developed system is around 78.4%. The overall character misclassification rate is observed as around 11%. Segmentation failures in the document pages account for around 10.6% error cases. The reason behind high segmentation failure is due to the over-segmentation of some of the constituent characters like 'i', 'j' and also due to under-segmentation of cursive words in the document pages. The designed system rejects around 9.24% characters in the test dataset. This is mainly due to the presence of multi-skewed handwritten text lines in the test documents. Completely cursive words were also rejected completely in many cases during the experimentation. Some of the sample word images successfully segmented and recognized by Tesseract are shown in Fig. 6(a-d). Fig. 7(a-b) shows some of the word images with erroneous segmentation and recognition results.

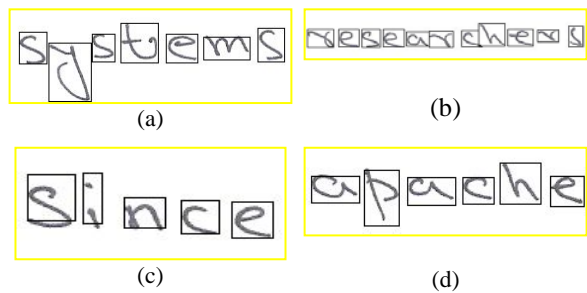

(a)  (b)

(c)  (d)

**Fig. 6(a-d). Some of the successfully segmented and recognized word images.**

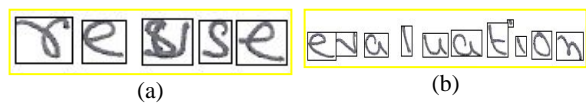

(a)  (b)

**Fig. 7. Some of the misclassified word images**
**(a) Recognition error in the 3$^{rd}$ character**
**(b) Internal segmentation in the 8$^{th}$ character**

## 5. Conclusion

As observed from the experimental results, Tesseract OCR engine fares reasonably with respect to the core recognition accuracy on user-specific handwritten samples of isolated / free-flow text, written using lower-case Roman script. The performance of the system is validated on a three-user recognition engine. A major drawback of the current technique is its failure to avoid over-segmentation in some of the characters. Also the system fails to segment cursive words in many cases leading to under-segmentation and rejection. The performance of the designed system may be improved by incorporating more training samples for each user and inclusion of word-level dictionary matching techniques.

## 6. Acknowledgements

One of the authors, Mr. Sandip Rakshit is thankful to the authorities of Techno India College of Technology for necessary supports during the research work. Dr. Subhadip Basu is thankful to the "Center for Microprocessor Application for Training Education and Research", "Project on Storage Retrieval and Understanding of Video for Multimedia" of Computer Science & Engineering Department, Jadavpur University, for providing infrastructure facilities during progress of the work.